\documentclass{article}
\usepackage{amsmath,amssymb}
\usepackage{subcaption}
\usepackage{microtype}
\usepackage{amsthm, amsmath}

\usepackage{bm}
\usepackage{graphicx}
\usepackage{natbib}
\usepackage[normalem]{ulem}

\newcommand{\EE}{\mathcal{E}}
\newcommand{\GG}{\mathcal{G}} 
\newcommand{\VV}{\mathcal{V}} 
\newcommand{\NN}{\mathcal{N}} 
\newcommand{\R}{\mathbb{R}}

\newcommand{\E}{\mathbb{E}}

\newcommand{\vv}{\mathbf{v}}
\newcommand{\ee}{\mathbf{e}}
\newcommand{\xx}{\mathbf{x}}
\newcommand{\yy}{\mathbf{y}}
\newcommand{\uu}{\mathbf{u}}
\newcommand{\zz}{\mathbf{z}}

\newcommand{\ie}{\emph{i.e.}}

\usepackage[accepted]{icml2021}

\icmltitlerunning{Data efficiency in graph networks through equivariance}

\begin{document}

\twocolumn[
\icmltitle{Data efficiency in graph networks through equivariance}

\icmlsetsymbol{equal}{*}

\begin{icmlauthorlist}
\icmlauthor{Francesco Farina}{to,equal}
\icmlauthor{Emma Slade}{to,equal}
\end{icmlauthorlist}

\icmlaffiliation{to}{GSK.ai, GlaxoSmithKline, London, N1C 4AG, UK}
\icmlcorrespondingauthor{Francesco Farina}{francesco.x.farina@gsk.com}
\icmlcorrespondingauthor{Emma Slade}{emma.x.slade@gsk.com}

\icmlkeywords{Machine Learning, ICML}

\vskip 0.3in
]

\printAffiliationsAndNotice{\icmlEqualContribution} 

\begin{abstract}
We introduce a novel  architecture for graph networks which is equivariant to any transformation in the coordinate embeddings that preserves the distance between neighbouring nodes. In particular, it is equivariant to the Euclidean and conformal orthogonal groups in $n$-dimensions. 
Thanks to its equivariance properties, the proposed model is extremely more data efficient with respect to classical graph architectures and also intrinsically equipped with a better inductive bias.
We show that, learning on a minimal amount of data, the architecture we propose can perfectly generalise to unseen data in a synthetic problem, while much more training data are required from a standard model to reach comparable performance. 
\end{abstract}

\section{Introduction}

The study of symmetries has been present for a long time in physics and mathematics. They are present in a wide variety of research fields and are a powerful tool for prediction and understanding systems on a deeper level. The mathematical framework which describes symmetries, group theory, has recently become attractive to researchers in machine learning for these same reasons~\citep{bronstein2021geometric}, with the additional aim of creating neural network architectures which are data efficient and less time-intensive to train.

By construction, graph neural networks~\citep{gori2005new,scarselli2008graph} (GNNs) are equivariant under the permutation group, which for appropriately structured datatypes, such as $n$-body systems, can greatly improve performance. It is therefore natural to consider whether, by considering other group transformations, we can construct an architecture which is more data efficient than a classic GNN.

There has been recent work on exploring group equivariance in graph networks, with particular on the Euclidean group. Equivariance to $SE(3)$ and $E(3)$ is obtained in~\citep{1802-08219,NEURIPS2020_15231a7c, kohler2020equivariant, finzi2020generalizing, batzner2021se3equivariant, Yang_2020_CVPR}. More generally, 
 an $E(n)$ equivariant message passing convolutional graph layer is proposed in~\citep{satorras2021en}, while in~\citep{horie2021isometric} equivariance to isometric transformations is considered.

In this work we define a novel graph architecture with modified update rules that make it equivariant to any transformation in the coordinate embeddings that preserves the distance between neighbouring nodes. We call the architecture a \emph{distance preserving graph network}, or DGN. In particular, we show that it is equivariant to the $n$-dimensional Euclidean group. An additional input layer makes it equivariant to scaling inputs via the conformal orthogonal group. Moreover, we show that the architecture is able to deal with affine transformations of the input coordinates.
  
Our network is able to, therefore,  filter out many copies of the same input transformed under these groups and consequently learns significantly more efficiently than one which considers the inputs as distinct such as a classic graph network. 
The architecture we propose therefore only requires a small subset of data points to train to high accuracy, as the equivariance properties enable it to recognise that a single sample contains the same information as many copies of it obtained by rotating, translating (and possibly scaling) it.

We demonstrate the potential of this type of architecture on a preliminary experiment by comparing it with a classical graph network on a classification task involving a synthetic dataset consisting of 3D geometric shapes.

\section{Group theory}
Symmetries of systems are described mathematically by group theory. 
A group can be defined as a set $G$ equipped with a binary operation, which enables one to combine two group elements to form a third, whilst preserving the group axioms (associativity, identity, closure and inverse).
Individual group elements correspond to individual symmetry operations, and the group as a whole encapsulates the full set of possible symmetry operations.

A mathematical object which is unaffected by a group operation is said to be \emph{invariant}, with invariance being a particular form of \emph{equivariance}. Formally, let $\varphi_g :X\to X$ be a transformation on $X$ for an abstract group $g \in G$. Then, the linear map $\Phi:X\to Y$ is \emph{equivariant} to  $G$ if $\forall \varphi_g: X \to X, \forall g \in G,$ $\exists\varphi_g ':Y\to Y$ such that
\begin{equation*}
 \Phi(\varphi_g(\xx)) = \varphi_g'(\Phi(\xx)) \,.
\end{equation*}
When $\varphi_g '$ is the identity, we say that $\Phi$ is invariant  to $G$.
Group equivariance is a powerful property and, whilst not often expressed in group theoretic notation, it is commonly employed by some deep learning architectures. For example, convolutional neural networks are equivariant under the translation group $T(n)$ and graph neural networks are invariant under the permutation group.

\subsection{$E(n)$ equivariance}\label{sec:construct_en_equiv}
Two groups of interest for physical systems are the translation $T(n)$ and rotation $O(n)$ groups. The semidirect product of these two groups $E(n) = T(n) \rtimes O(n)$ is known as the Euclidean group.
The Euclidean group encapsulates symmetries which may exist in large amounts of physical data, such as an image (or representation of an image) which may have been rotated or moved with respect to the centre.
A map $\varphi$ is $E(n)$ equivariant if $\varphi: \E^n \to \E^n$, where $\E^n$ is a Euclidean space. As an example of a Euclidean space one can consider $\R^n$ equipped with an inner product (assuming an inner product to exist without loss of generality).
To find a transformation that is $E(n)$ equivariant, note that, a translation affects a point $\xx$ as
  \begin{equation*}
    \xx \to \xx^+= \xx + \zz, \quad \zz \in \E^n \,,
  \end{equation*}
while a rotation as
  \begin{equation*}
    \xx \to \xx^+= Q\xx, \quad Q \in O(n) \,.
  \end{equation*}
Now, by using the above relations and recalling that $E(n)$ is the semidirect product of $T(n)$ and $O(n)$, it is easy to show that the function $\|\xx_i-\xx_j\|_2^2$ is equivariant under $E(n)$ since
\begin{subequations}\label{eq:En_eq}
\begin{align}
  \|\xx_i - \xx_j\|_2^2 &\to \|\xx_i^+ - \xx_j^+\|_2^2 \\
  &= \|Q(\xx_i + \zz) - Q(\xx_j + \zz) \|_2^2\\
    &= \|Q\xx_i - Q\xx_j  \|_2^2 \\
    &= (\xx_i - \xx_j  )^\top Q^\top Q (\xx_i  - \xx_j ) \\
    &= \|\xx_i - \xx_j  \|_2^2 \,,
\end{align}
where we used the fact that $Q^\top Q=I$ for all $Q \in O(n)$.
\end{subequations}

\subsection{Conformal orthogonal equivariance} 

The conformal orthogonal group is the group of dilations and orthogonal rotations. It is a subgroup of the conformal group, which preserves only angles and contains additional transformations such as inversions, which we do not wish to preserve as they do not preserve distances between the coordinates $\xx$.
Formally, the conformal orthogonal group $\text{CO}(V,\mathcal{Q})$ is defined for a vector space $V$ with a quadratic form $\mathcal{Q}$. It contains the linear transformations $\varphi : \mathcal{T} \to V$ and its action is defined as
\begin{equation} \label{eq:conformal_group}
\mathcal{Q}(\mathcal{T}x) = \gamma^2 \mathcal{Q}(x) \,,
\end{equation}
where $\mathcal{T}$ is the set of linear transformations that we need to define and $\gamma$ is a scalar.
As we restrict our discussion to Euclidean spaces, in particular $\E^n$, the quadratic form to consider is
\begin{equation} \label{eq:euclidean_distance}
\mathcal{Q} = \sum_i x_i^2 \,.
\end{equation}
By using~\eqref{eq:euclidean_distance} in~\eqref{eq:conformal_group} we have the condition
\begin{equation*}
\sum_i (\mathcal{T}  x_i)^2 = \gamma^2 \sum_i x_i^2 \,.
\end{equation*}
 If we consider, as above, an orthogonal transformation $Q$, and in addition, a dilation $x_i \to \gamma x_i$, then, with $\mathcal{T} = \gamma Q$
\begin{align*}
\sum_i (\gamma Q  x_i)^2 &= \gamma^2 Q^\top Q \sum_i x_i^2 \\
 &= \gamma^2  \sum_i x_i^2 \,,
\end{align*}
where we again used the orthogonality of $Q$.
Therefore, with a suitable definition of $\gamma$ and $\xx$, we can construct an architecture equivariant under~\eqref{eq:conformal_group}.
In addition, as shown in Sec.~\ref{sec:construct_en_equiv}, we can also construct an architecture which is equivariant under $E(n)$, if we consider $\|\xx_i-\xx_j\|_2^2$. In all, with the group definitions for $E(n)$ and $CO(\R^n, \mathcal{Q})$ now defined, and the actions of a coordinate system under these groups, we can build an architecture which can take advantage of these group properties. Such an architecture will be able to deal with rotated, translated and scaled data more efficiently than a standard graph network which does not have the above group properties built into it and considers all data to be distinct.

\section{Toward equivariant graph networks}
In order to build a graph network with these additional group structures, we start by introducing general graph networks. Then, we build on the notion of equivariance to redefine the network updates to be $E(n)$ equivariant and obtain scale equivariance via a rescaling procedure for the input.

\subsection{Graph networks}
A graph is defined as $\GG(\VV,\EE)$ where $\VV=\{1,\dots, N\}$ is the set of nodes and $\EE=\{(j,i)\}\subseteq \VV\times\VV$ is the set of (directed) edges connecting nodes in $\VV$ (where $j$ and $i$ denote the source and target nodes respectively). We define $\NN_i\triangleq\{j\mid (j, i)\in\EE\}$ as the set of (in-)neighbours of node $i$. Moreover, we associate feature node and edge embeddings, $\vv_i\in\R^{n_v}$, $\forall i\in\VV$, and $\ee_{ji}\in\R^{n_e}$, $\forall (j,i)\in\EE$ respectively, to each node and edge in the graph as long as a global attribute $\uu\in\R^{n_u}$.
Then, a general standard graph network block can be defined in terms of edge, node and global updates as
\begin{subequations}\label{eq:gnn}
\begin{align}
  \ee_{ji}^+ &= \phi^e\big(\vv_j, \vv_i, \ee_{ji}, \uu\big),\, \forall (j,i)\in\EE\\
  \vv_i^+ &= \phi^v\big(\vv_i, \rho^{e\to v}\big(\{\ee_{ji}^+\}_{j\in\NN_i}\big), \uu\big),\,\forall i\in\VV\\
  \uu^+ &= \phi^u\big(\rho^{v\to u}(\{\vv_{i}^+\}_{i\in\VV}), \rho^{e\to u}(\{\ee_{ji}^+\}_{(j,i)\in\EE}), \uu\big)
\end{align}
\end{subequations}
where $\phi^e:\R^{2n_v + n_e+n_u}\to \R^{n_e^+}$, $\phi^v:\R^{n_v+ n_e^+ + n_u}\to \R^{n_v^+}$, $\phi^u:\R^{n_v^+ + n_e^+ + n_u}\to \R^{n_u^+}$ are update functions (usually defined as neural networks whose parameters are to be learned) and $\rho^{e\to v}, \rho^{e\to u}, \rho^{v\to u}$ are aggregation functions reducing a set of elements of variable length to a single one via an input's permutation invariant operation like element-wise summation, mean or maximum.

\subsection{Distance preserving graph network block (DGN)}
With a slight abuse of notation, let us define $\GG_X(\VV,\EE)$ as a graph embedding where, in addition to the feature node embeddings $\vv_i\in\R^{n_v}$, coordinate embeddings $\xx_i\in\R^{n_x}$ are also associated to each node $i\in\VV$. 

If we assume that the initial node embeddings $\vv_i$ contain no absolute coordinate or orientation information about the initial coordinate embeddings $\xx_i$ then the update of our equivariant graph network is given by 
 \begin{subequations}\label{eq:En}
  \begin{alignat}{3}
    &\ee_{ji}^+ &&= \phi^e\big(\ee_{ji}, \vv_i, \vv_j, \|\xx_i-\xx_j\|_2^2, \uu\big),\,\forall (j,i)\in\EE \label{eq:En_edge_update} \\
    &\vv_i^+ &&= \phi^v\big(\rho^{e\to v}\big(\{\ee_{ji}^+\}_{j\in\NN_i}\big), \vv_i, \uu\big),\,\forall i\in\VV\label{eq:En_node_update}\\
    &\xx_i^+ &&= \psi^x\left(i,\GG_X\right),\,\forall i\in\VV \label{eq:En_coord_update}\\
    &\uu^+ &&= \phi^u\big(\rho^{e\to u}(\{\ee_{ji}^+\}_{(j,i)\in\EE}),\rho^{v\to u}(\{\vv_{i}^+\}_{i\in\VV}),\nonumber \\ & && \rho^{x\to u}(\{\|\xx_{i}^+-\xx_j^+\|_2^2\}_{(j,i)\in\EE}),\uu\big)\label{eq:En_global_update}
  \end{alignat}
\end{subequations}
where $\phi^e:\R^{n_e+ 2n_v+ n_u+1}\to \R^{n_e^+}$, $\phi^v:\R^{n_e^+ + n_v+ n_u}\to \R^{n_v^+}$, $\phi^u:\R^{n_e^+ + n_v^+ + n_u+1}\to \R^{n_u^+ }$ are the update functions to be learned, $\rho^{e\to v}, \rho^{e\to u}, \rho^{v\to u}$ are aggregation functions and $\psi^x:\R^K\to \R^{n_x}$ is some, possibly parametric, relative distance preserving map. Formally, let $\GG_X(\VV,\EE)$ and $\GG_Y(\VV,\EE)$ be two different coordinate embeddings of the same graph such that $\|\xx_i-\xx_j\|^2 = \|\yy_i-\yy_j\|^2$, $\forall (i,j)\in\EE$. Let $\xx_i^+=\psi(i, \GG_X)$ and $\yy_i^+=\psi(i,\GG_Y)$ for some function $\psi:\R^K\to \R^{n}$, $K\in\R^+$. We say that $\psi$ is a \emph{relative distance preserving} (RDP) map if $\|\xx_i^+-\xx_j^+\|^2 = \|\yy_i^+-\yy_j^+\|^2$, $\forall (i,j)\in\EE$.
Different structures can be imposed on $\psi$, possibly based on what one wants to achieve. Examples include the identity function $\xx_i^+=\xx_i$ or any transformation that modifies all the coordinates in the same way, like $\xx_i^+=A\xx_i+q$, with $A$ and $q$ possibly being parametric functions.
If one further assumes $\yy_i$ is a Euclidean transformation of $\xx_i$, more complex relative distance preserving maps can be obtained, like
\begin{equation}
  \xx_i^+ = \xx_i + \sum_{j \in \NN_i} a_{ji}(\xx_j-\xx_i).\label{eq:eq_map}
\end{equation}
with $a_{ji}$ possibly being a parametric function of node/edge/global attributes.

\begin{figure}
  \centering
  \includegraphics[width=0.7\columnwidth]{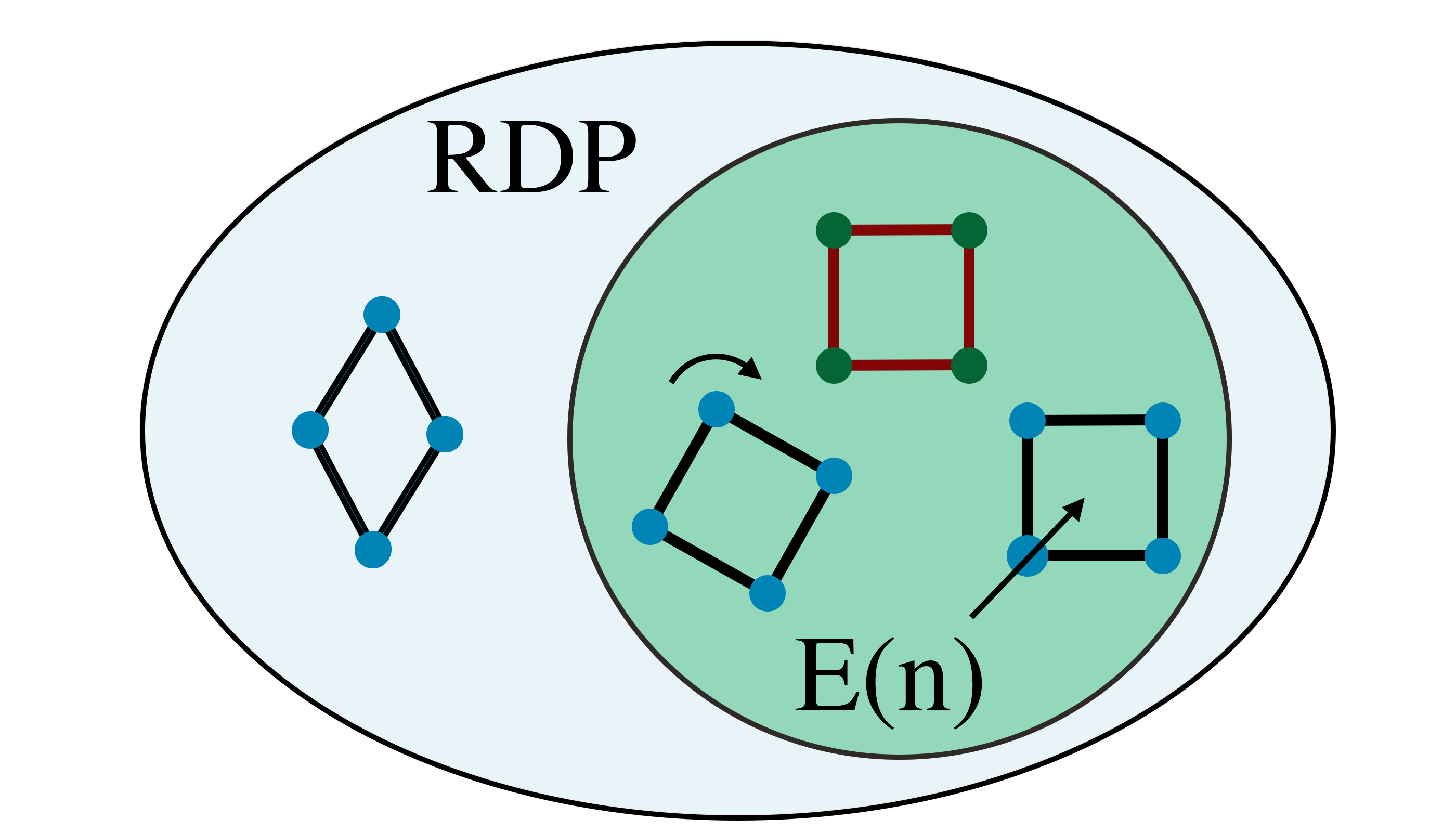}
  \caption{Graphical representation of transformations in the coordinate embeddings to which the DGN block is equivariant.}
  \label{fig:diagram}
\end{figure}

\paragraph{Equivariance properties}
Thanks to its update structure~\eqref{eq:En}, the DGN block is, by construction, equivariant under any relative distance preserving transformation of the coordinate embeddings. In particular, it is equivariant to Euclidean transformations.
To show this, let us start by showing that the edge update,~\eqref{eq:En_edge_update} is equivariant under $E(n)$. Thanks to ~\eqref{eq:En_eq} we know that $ \|\xx_i-\xx_j\|_2^2$ is equivariant under an $E(n)$ transformation, and we can easily show
\begin{align*}
 \phi^e\big(\ee_{ji}, \vv_i, & \vv_j, \|\xx_i-\xx_j\|_2^2, \uu\big)  \\ &\to \phi^e\big(\ee_{ji}, \vv_i, \vv_j, \|Q\xx_i + \zz -Q \xx_j - \zz\|_2^2, \uu\big) \\
&=   \phi^e\big(\ee_{ji}, \vv_i, \vv_j, \|\xx_i - \xx_j \|_2^2, \uu\big) \,.
\end{align*}
As for the coordinate update~\eqref{eq:En_coord_update} it is trivially equivariant under $E(n)$ thanks to $\psi^x$ and the fact that the coordinates are affected by Euclidean transformations.
Finally, the equivariance of the node and global updates, ~\eqref{eq:En_node_update} and~\eqref{eq:En_global_update}, follow naturally, being themselves functions of equivariant quantities. Figure~\ref{fig:diagram} pictorially shows transformations to which the DGN is equivariant for a 2D graph.

\subsection{Scale invariance through input scaling}
Whilst equivariance to the $E(n)$ group can be achieved via modified update rules, building equivariance to the conformal orthogonal group (and hence scaling) requires rescaling the information passed to the network. Let
\begin{equation}
  \gamma = \frac{\alpha}{\max\limits_{(i,j)\in\EE} \|\xx_i-\xx_j\|} \label{eq:coordinate_scaling}
\end{equation}
for some $\alpha\in \R$, where we align $\gamma$ as defined in ~\eqref{eq:coordinate_scaling} with the one defined in ~\eqref{eq:conformal_group}. Then, conformal orthogonal invariance can simply be obtained by computing scale-normalised coordinates $\tilde{\xx}_i=\gamma \xx_i$ and using them as the coordinates in~\eqref{eq:En}. This input scaling layer can be applied to the input of a DGN to obtain equivariance of the coordinate embeddings to translations, rotations and dilations.

\section{Experimental results}

In this section, we consider a classification problem with a dataset composed of graph representations of $3$D polytopes consisting of simplexes, cubes, octahedra, dodecahedra and icosahedra. Each vertex and edge of the polytopes correspond to a node and and edge in the graph, so that each polytope is specified in terms of node coordinates and a list of edges. The training dataset consists of a single graph per polytope (hence $5$ graphs in total) while the test set is composed of $500$ randomly transformed versions of those in the training set with $100$ per class. Coordinates for the polytopes in the test set are obtained as $\tilde{\xx}_i\! =\! \gamma A\xx_i + q$ for some $\gamma\in\R$, $A\in\R^{3\times 3}$ and $q\in\R^3$, where $\xx_i$ are the coordinates of the graphs in the training set. We consider different transformations corresponding to different choices of $\gamma, A, q$.

We compare graph networks built with DGN and standard blocks. The DGN block is also combined with a scaling layer (SDGN) to obtain equivariance to dilations. 

Each network is composed of $3$ graph layers after which the produced node embeddings are passed through a MLP, a global (sum) pooling layer and a final MLP. The output dimension is equal to the number of classes for graph classification tasks, \ie, $5$. 
The update functions of each graph layer are implemented as MLPs. In particular the coordinate update is implemented as the identity function $\xx_i^+=\xx_i$.
All the MLPs consist of a single hidden layer containing $64$ neurons and swish activation function. The hidden graph layers have dimension $32$.

Adam~\citep{kingma2014method} is used to train the all the models with a learning rate $\alpha=0.001$, no regularisation and batch-size equal to the number of training samples ($5$) for $500$ epochs. To obtain errors on the results, each network is trained $10$ times, with differing initialisations.

\begin{figure*}[t!]
  \centering
  \begin{subfigure}[t]{0.19\linewidth}
    \centering
    \includegraphics[height=0.7\linewidth]{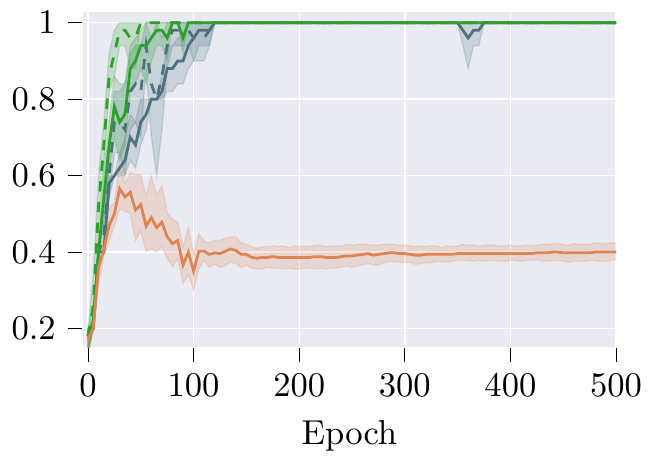}
    \caption{Orthogonal}
  \label{fig:shapes_TO}
  \end{subfigure}
  \begin{subfigure}[t]{0.19\linewidth}
    \centering
    \includegraphics[height=0.7\linewidth]{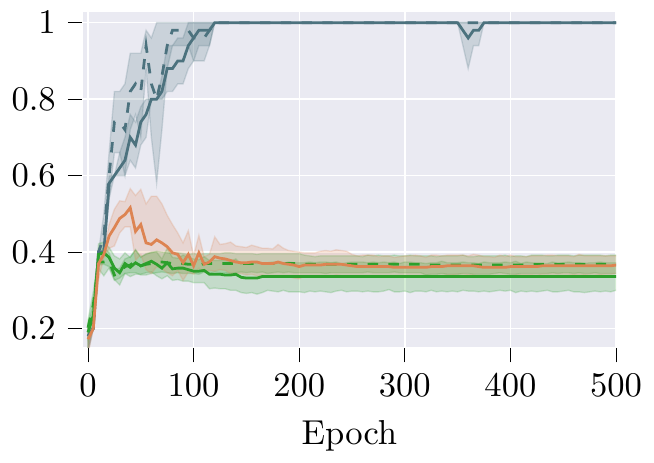}
    \caption{Orthogonal + dilation}
  \label{fig:shapes_TOS}
  \end{subfigure}
  \begin{subfigure}[t]{0.19\linewidth}
    \centering
    \includegraphics[height=0.7\linewidth]{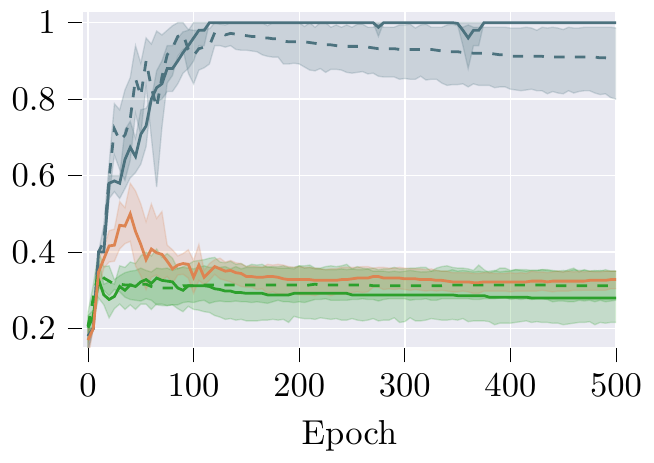}
    \caption{Non-orthogonal ($\mu=0.5$)}
  \label{fig:shapes_TOS_affine}
  \end{subfigure}
  \begin{subfigure}[t]{0.19\linewidth}
    \centering
    \includegraphics[height=0.7\linewidth]{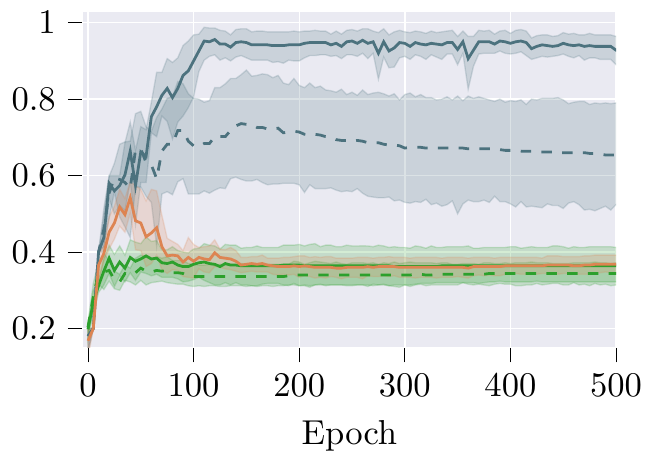}
    \caption{Non-orthogonal ($\mu=1.5$)}
  \label{fig:shapes_TOS_affine_mid}
  \end{subfigure}
  \begin{subfigure}[t]{0.19\linewidth}
    \centering
    \includegraphics[height=0.7\linewidth]{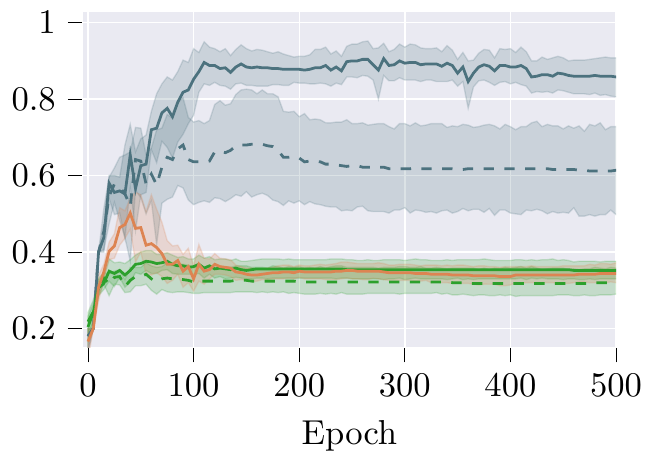}
    \caption{Non-orthogonal ($\mu=3$)}
  \label{fig:shapes_TOS_affine_high}
  \end{subfigure}
  \includegraphics[width=0.6\linewidth]{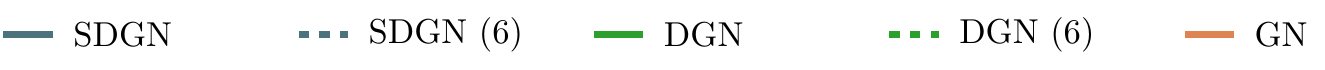}
  \caption{Polytopes classification: test accuracy for the different architectures on the different test sets.}
  \label{fig}
\end{figure*}

Next we describe in detail the different experiments we performed by considering different transformations in the test set. Results are summarised in Table~\ref{tab:table_res} in terms of accuracy on the test set (mean and standard deviation) and in Figure~\ref{fig} in terms of convergence rate (mean and min-max range). As we see, while all networks obtain a perfect training accuracy, this is in stark contrast to when they are presented with unseen data.

\begin{table}
  \centering
  \resizebox{\columnwidth}{!}{
  \begin{tabular}{l|c|c|c|c|c|c|c|}
    & & & \multicolumn{5}{|c|}{test accuracy}  \\\cline{4-8}
    block &$\psi^x$  & train acc & Orthogonal  & Orthogonal + dilation  & \begin{tabular}{c}Non-orthogonal\\($\mu=0.5$) \end{tabular} & \begin{tabular}{c}Non-orthogonal\\($\mu=1.5$) \end{tabular} & \begin{tabular}{c}Non-orthogonal\\($\mu=3.0$) \end{tabular}\\
    \hline\hline
    SDGN & I      & $1$   & $\bm{1}$        & $\bm{1}$        & $\bm{1}$       & $\bm{0.91\pm 0.07}$ & $\bm{0.83\pm 0.11}$\\
    SDGN & \eqref{eq:eq_map}      & $1$   & $\bm{1}$            & $\bm{1}$        & $\bm{0.89\pm 0.16}$ & $\bm{0.62\pm 0.24}$ & $\bm{0.60\pm 0.21}$\\
    DGN & I      & $1$   & $\bm{1}$        & $0.34\pm 0.05$  & $0.28\pm 0.12$ & $0.36\pm 0.07$ & $0.35\pm 0.04$ \\
    DGN & \eqref{eq:eq_map}       & $1$   & $\bm{1}$            & $0.37\pm 0.04$  & $0.31\pm 0.06$ & $0.34\pm 0.04$ & $0.32\pm 0.05$ \\
    GN    & -    & $1$   & $0.40\pm 0.05$  & $0.36\pm 0.04$  & $0.32\pm 0.05$ & $0.38\pm 0.07$ & $0.35\pm 0.05$ \\
    \hline
  \end{tabular}
  }
  \caption{Training and test accuracy (mean $\pm$ standard deviation over $10$ runs) for different transformations in the test set. }
  \label{tab:table_res}
\end{table}

\paragraph{Orthogonal transformations}
We start by considering only orthogonal rotations and translations in the test set (\ie, $\gamma=1$, $A^\top A=1$).
Rotation angles and shifts are drawn from a Gaussian distribution of appropriate size with $0$ mean and unit variance. As expected both DGN and SDGN are able to perfectly generalise to the test set thanks to their equivariance to rotations and translations, while the standard GNN performs poorly.

\paragraph{Orthogonal transformations and dilations}
Now we consider the case where in addition to orthogonal rotations ($A^\top A=1$) and translations, dilations are also added to the test set ($\gamma\in\R$). The scale factor $\gamma$ is drawn from a Gaussian distribution with unit mean and standard deviation. Now, due to dilations, the DGN (as the GNN) is unable to properly cope with graphs in the test dataset; only the SDGN can perfectly generalise to unseen data. 

\paragraph{Non-orthogonal transformations}
Finally, we consider the case of random non-orthogonal transformations in the test set (\ie, $\mu = \text{E}[\|A^\top A - I\|_F]>0$, $\gamma\in\R$). 
In this case, the non-orthogonality of a rotation matrix breaks the equivariance under $E(n)$, instead belonging to the larger group of affine transformations which allows shearing and squashing of the data. 
These type of transformations are not explicitly taken into account by the DGN block. However, we see from Table~\ref{tab:table_res} that, on this specific dataset, the SDGN is able to generalise well also to non-orthogonally transformed data. Thus, it seems that thanks to its equivariance properties, our architecture is able to learn affine transformations quite well from few data points.

\begin{figure}
  \centering
  \includegraphics[width=\columnwidth]{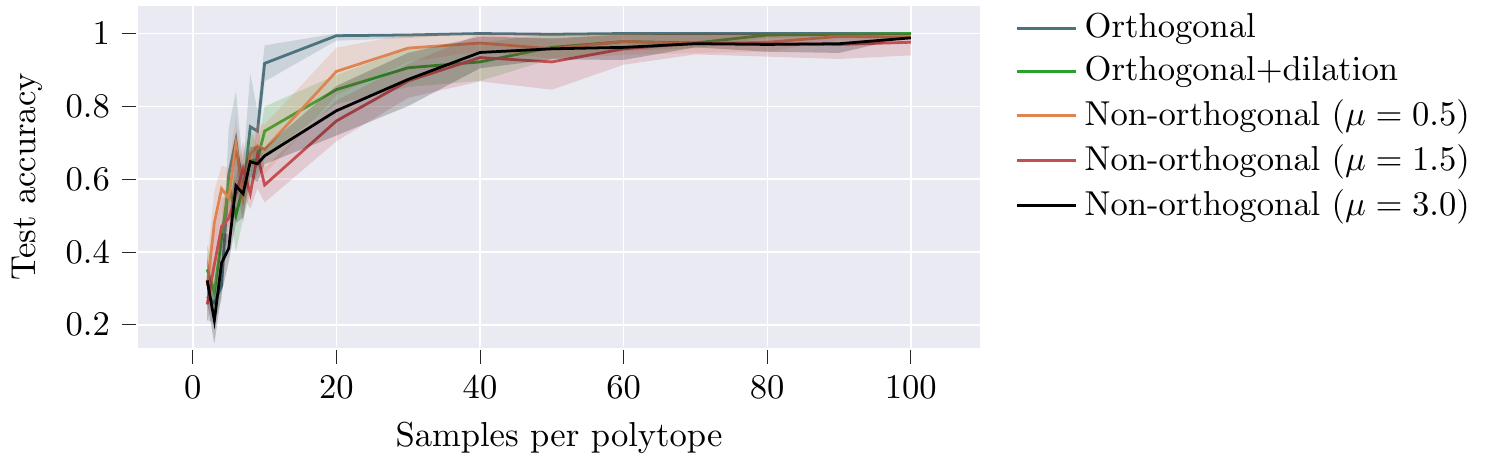}
  \caption{Test accuracy vs samples per polytope in the training set for a standard GNN.}
  \label{fig:efficiency}
\end{figure}

\paragraph{Data efficiency}
To emphasise the advantage of having a network that is able to exploit symmetries in the dataset in terms of data efficiency, we study how many samples in the training set are necessary for a standard GNN to reach reasonable generalisation performance. For the set of transformations we considered in the previous sections, we augmented the training set with $\{2,3,\dots,100\}$ randomly transformed (as in the respective test set) copies of each polytope. We trained a standard GNN on these augmented datasets and observed the resulting test accuracy after $500$ epochs.
Results are reported in Figure~\ref{fig:efficiency} for each set of transformations in terms of mean and standard deviation over $10$ random initialisations. It can be seen that $20$ samples per polytope may be sufficient when only orthogonal transformations are considered. Adding also dilations and non-orthogonal transformations further increases the number of data points that are required.
This shows that while data augmentation can be successfully exploited, it is provably sub-optimal in terms of sample complexity~\citep{mei2021learning} and architectures with built-in equivariance properties represent a more efficient strategy to consider.

\section{Conclusion}
In this paper we have outlined a novel deep learning architecture which is equivariant to the $E(n)$ and conformal orthogonal groups, with additional equivariance to the permutation group.
As a preliminary application of the architecture, we have applied our model to a custom dataset composed of $3$D geometric shapes. We have shown that the architecture we propose is significantly more accurate than a standard graph network on data which has been shifted, rotated and scaled. Furthermore, even in  the presence of non-orthogonal transformations, the proposed architecture seems to outperform standard graph networks when learning with few data points. Indeed, the model is significantly more data efficient than a standard graph network and able to train with fewer training points, as it does not consider group-transformed graphs as distinct and can perfectly generalise to properly transformed ones. Whilst we have shown that our architecture is able to train with significantly less data than a classic GNN, it would be interesting to determine the optimal subset of data for the DGN architecture from a large dataset with symmetries embedded therein. In doing so, we would be able to reduce even further the data requirements of the architecture, at no expense to its accuracy.

\bibliographystyle{plainnat}
\bibliography{biblio.bib}

\end{document}